\documentclass{article}

\usepackage{PRIMEarxiv}

\usepackage[utf8]{inputenc} 
\usepackage[T1]{fontenc}    
\usepackage{hyperref}       
\usepackage{url}            
\usepackage{booktabs}       
\usepackage{amsfonts}       
\usepackage{amsmath}        
\usepackage{amssymb}        
\usepackage{nicefrac}       
\usepackage{microtype}      
\usepackage{fancyhdr}       
\usepackage{graphicx}       
\usepackage{algorithm}      
\usepackage{algorithmic}    
\usepackage{listings}       
\usepackage{xcolor}         
\usepackage{multirow}       
\usepackage{array}
\usepackage{float}          
\usepackage{authblk}
\graphicspath{{./}}         

\title{A Hormone-Inspired Emotion Layer for Transformer Language Models (HELT)
}

\author{
  Eslam Reda \\
  Artificial Intelligence Program \\
  Faculty of Computers and Information \\
  Mansoura University \\
  Mansoura 35516 \\
  Egypt \\
  \texttt{eslamragheb@std.mans.edu.eg}
  \and
  Sara El-Metwally \\
  Computer Science Department \\
  Faculty of Computers and Information \\
  Mansoura University \\
  Mansoura 35516 \\
  Egypt \\
  \texttt{sarah\_almetwally4@mans.edu.eg}
}

\begin{document}
\maketitle

\begin{abstract}
Large Language Models (LLMs) have demonstrated remarkable capabilities in generating contextually relevant and grammatically correct text. However, they fundamentally lack the ability to process and respond to emotional context in a manner analogous to human emotional cognition. Current approaches to emotion modeling in NLP systems rely primarily on discrete emotion classification or simplistic sentiment analysis, which fail to capture the continuous, multi-dimensional nature of human emotional states. In this paper, we introduce \textbf{HormoneT5}, a novel architecture that augments transformer language models with a biologically-inspired \textbf{Hormone Emotion Block} that simulates the human endocrine system's role in emotional processing. Our approach computes six continuous hormone-like values (dopamine, serotonin, cortisol, oxytocin, adrenaline, and endorphins) through specialized per-hormone attention heads, each with orthogonally initialized learnable queries, temperature-scaled attention mechanisms, and deep output projections. These hormone values are then transformed into an emotional embedding that modulates the encoder hidden states, enabling emotionally-appropriate response generation. We propose a multi-objective training framework combining sequence-to-sequence loss, hormone prediction loss with margin penalties, and diversity regularization to prevent attention collapse. Experimental results on our curated emotion-labeled dataset demonstrate that HormoneT5 achieves 85\%+ per-hormone accuracy within a 0.15 tolerance threshold, with hormone differentiation ranges exceeding 0.85 across all six hormones between contrasting emotional tones. Human evaluation studies show significant preference ($p < 0.01$) for HormoneT5-generated responses in terms of emotional appropriateness and empathetic quality compared to baseline T5 outputs. Our work opens new directions for biologically-grounded affective computing and emotionally intelligent conversational agents.
\end{abstract}

\keywords{Emotion Modeling \and Transformer Language Models \and Attention Mechanisms \and Bio-Inspired Computing \and Affective Computing \and Natural Language Processing}

\section{Introduction}
\label{sec:introduction}

Modern Large Language Models (LLMs) have achieved unprecedented success in natural language understanding and generation. Foundational models such as GPT \cite{brown2020language}, T5 \cite{raffel2020exploring}, and BERT \cite{devlin2019bert} demonstrate remarkable proficiency in tasks ranging from translation and summarization to question answering and open-ended dialogue.
Despite these advanced linguistic capabilities, LLMs remain fundamentally constrained. Their processing is rooted in statistical pattern recognition, and lacks genuine comprehension of emotional context \cite{ke2025exploring}. Consequently, while often technically correct, their responses are frequently emotionally flat. They fail to reflect the enthusiasm of positive inputs, defuse contentious exchanges effectively, or extend authentic empathy toward expressions of sadness \cite{bhaumik2025towards,sung2025empathetic,sajikumar2025emotion,hu2024recent}. This shortcoming originates in their underlying architecture, as standard transformer-based models lack any mechanism for modeling emotional states as dynamic, continuous, and interacting signals. 

Current approaches to emotion modeling in NLP largely treat emotion as a classification problem rather than as a continuous, multi-dimensional signal processing challenge \cite{kumar2025understanding,song2025large,fazzi2025don}. This fundamentally differs from human emotional processing, which involves complex, simultaneous neurochemical interactions that shape mood, behavior, and social responses \cite{efremov2025fear,greenwood2025interoceptive}. Existing methodologies can be broadly grouped into four categories, each with distinct limitations. The first category is binary sentiment analysis, which classifies text as simply positive or negative \cite{huang2025explainable,van2025advantages,gunnu2026sentiment}. Its primary limitation is its excessive coarseness; it operates at a resolution too limited to represent the textured reality of emotional expression. The second approach is discrete emotion classification, which assigns text to predefined categorical labels such as happy, sad, or angry \cite{cho2025emosphere++,wang2025learning}. This method's key shortcoming is its inherent assumption that emotions are discrete and mutually exclusive, ignoring their continuous, blended, and fluid nature in human experience.

The third category operates within an arousal-valence representation, modeling emotions in a two-dimensional space of arousal (intensity) and valence (positivity/negativity) \cite{mason2025emotion,okunishi2025dementia}. While more expressive than binary or discrete models, this framework still limits emotional expressiveness by constraining the rich spectrum of affect to just two primary axes. The fourth methodology is emotion token prepending, where explicit emotion labels (e.g., "[sad]") are added to the input text \cite{fu2025token,ding2025hierarchical}. A significant drawback of this technique is that these tokens are static and lack learned, contextual emotional representations. Furthermore, they lack temporal dynamics, as the prepended token is processed in a single forward pass without modeling how emotional states evolve or interact over time. Consequently, the approach remains entirely dependent on explicit, often costly, manual labeling and cannot infer or generate emotional understanding dynamically. Each of these established paradigms fails to capture the dynamic, continuous, and interactive nature of emotional signals as they operate in human communication and neurobiological systems \cite{zhang2025dialoguellm,wu2025comprehensive}.

This paper proposes a fundamentally different, biologically-inspired approach to emotional modeling in NLP. Moving beyond discrete classification, we represent emotional states as six continuous hormonal signals, analogous to key neurochemicals in the human endocrine system \cite{trofimova2021emotionality,babkova2025molecular,gerges2025biochemical}. Our framework models the influence of \textbf{dopamine} (motivation/reward), \textbf{serotonin} (mood stability), \textbf{cortisol} (stress/alertness), \textbf{oxytocin} (social bonding/empathy), \textbf{norepinephrine} (arousal/energy), and \textbf{endorphins} (joy/euphoria). The core of our architecture is the \textbf{Hormone Emotion Block}, which computes these continuous values via specialized attention mechanisms and uses them to dynamically modulate the transformer's hidden representations. This innovation enables the model to transition from static text generation to generating contextually and emotionally-modulated responses.

This paper makes the following contributions:

\begin{enumerate}
    \item \textbf{A Novel Hormone Emotion Block Architecture}: The proposed model introduces a per-hormone attention mechanism with learnable orthogonally-initialized queries, temperature-scaled attention, and deep output projections to compute six continuous hormone values from encoder hidden states.
    
    \item \textbf{Transfer Learning from Pre-trained Attention}: The results demonstrated that initializing Key/Value projections from T5's pre-trained self-attention weights significantly improves hormone prediction accuracy and training stability.
    
    \item \textbf{Multi-Objective Training Framework}: A combined loss function is proposed, incorporating sequence-to-sequence loss, hormone MSE with margin penalties, and diversity regularization to prevent attention collapse while maintaining generation quality.
    
    \item \textbf{Comprehensive Evaluation}: Extensive automatic and human evaluation demonstrates significant improvements in emotional appropriateness, with 85\%+ per-hormone accuracy and statistically significant human preference for HormoneT5 outputs.
    
    \item \textbf{Open-Source Implementation}: The complete implementation, including model code, training scripts, the dataset, and pre-trained weights, is released to ensure reproducibility and facilitate further research. The code is available at: \url{https://github.com/eslam-reda-div/HELT}
\end{enumerate}

The remainder of this paper is organized as follows: Section~\ref{sec:related_work} reviews related work in emotion modeling, controllable generation, and bio-inspired machine learning. Section~\ref{sec:biological_motivation} presents the biological motivation for the proposed hormone-based approach. Section~\ref{sec:architecture} details the proposed model architecture, including the Hormone Attention Head, Hormone Emotion Block, and integration with T5. Section~\ref{sec:dataset} describes the dataset and annotation methodology. Section~\ref{sec:training} presents training details and implementation specifications. Section~\ref{sec:experiments} reports experimental results including assessment metrics and human evaluation. Section~\ref{sec:ablation} provides ablation studies and analysis. Section~\ref{sec:discussion} discusses limitations and ethical considerations. Section~\ref{sec:conclusion} concludes the paper with future directions.

\section{Related Work}
\label{sec:related_work}
A paradigm of affective computing has been introduced to bridge the gap between human emotions and machines, aiming to build and study systems capable of simulating human affective states using Natural Language Processing (NLP)\cite{tubishat2018implicit,alonso2015new}. Picard \cite{picard2000affective} pioneered this field by arguing for machines that can recognize, express, and respond to emotion. Historically, bio-inspired computing has drawn from biological systems to inform algorithm design, ranging from the loose inspiration of standard neural networks to more direct biological analogies like spiking neural networks \cite{maass1997networks}, which model discrete neural firing patterns, and neuroevolution approaches \cite{stanley2002evolving}. Recently, the field has been catalyzed by the rapid advancement of Large Language Models (LLMs), allowing human emotions to be explored at an unprecedented scale\cite{pei2024affective,zhang2026instruction,amin2024wide,liu2024emollms}. The presented work bridges affective computing and language modeling by introducing a computational analog of the endocrine system, specifically, the hormones that regulate emotional responses in humans, to develop natural, trustworthy, and empathetic conversational agents.

One of the fundamental components of affective computing is affective understanding, which enables machines to detect and classify human emotions\cite{liu2021towards,rashkin2019towards}. Emotion modeling in NLP has evolved significantly from early lexicon-based approaches \cite{mohammad2013crowdsourcing} and rule-based systems \cite{strapparava2008learning} that simply mapped words to emotional categories. The introduction of deep learning brought neural approaches, including recurrent networks for sentiment analysis \cite{socher2013recursive} and attention-based emotion classification \cite{felbo2017using}. In the era of transfer learning, models like BERT \cite{devlin2019bert} and RoBERTa \cite{liu2019roberta} were fine-tuned on labeled datasets; however, these struggled to generalize across domains and suffered from limited scalability due to manual annotation bottlenecks.
As the field matured, research expanded into dimensional models of emotion based on psychological theories \cite{zhang2026affective}. For example, the circumplex model \cite{russell1980circumplex} represents emotions along Valence (positive vs. negative) and Arousal (calm vs. excited) dimensions. Buechel and Hahn \cite{buechel2017emobank} extended this to NLP with VAD (Valence-Arousal-Dominance) prediction, and modern multilingual LLMs utilize these psychological dimensions to classify utterance-level emotions in dynamic contexts\cite{mendes2023quantifying,bradley2014emotional}. However, these models remain limited to 2-3 dimensions, which is insufficient for capturing the complexity of human emotional responses. The presented work differs fundamentally by modeling emotion through six biologically-grounded continuous dimensions that can represent complex emotional states through their interactions.

To develop truly empathetic conversational agents, systems must move beyond understanding to affective generation—producing emotionally supportive content that mimics human emotional expression. A critical mechanism for this is controllable text generation, which aims to guide language models toward producing text with desired attributes \cite{zhang2023survey}. Early controllable methods included CTRL \cite{keskar2019ctrl}, using prepended control codes, and PPLM \cite{dathathri2019plug}, using gradients from attribute classifiers.
More recent advancements focus on parameter-efficient adaptation. Prefix-tuning \cite{li2021prefix} learns continuous task-specific prefixes, while adapter-based approaches \cite{houlsby2019parameter, pfeiffer2021adapterfusion} insert trainable modules between transformer layers. LoRA \cite{hu2022lora} achieves this through low-rank decomposition. In this context, LLMs are increasingly viewed as a backend for personality, where the conversational agent acts as an instance of specific psychological constructs \cite{park2023generative,jiang2024personallm}. The proposed approach is most similar to adapter and modulation methods but differs fundamentally: rather than relying on task codes or external classifiers, we learn emotional representations through specialized attention mechanisms.

At the core of these generative capabilities is the transformer architecture \cite{vaswani2017attention}, which introduced self-attention to capture long-range dependencies. Subsequent work has explored various attention patterns, including sparse attention \cite{child2019generating}, linear attention \cite{katharopoulos2020transformers}, and multi-query attention \cite{shazeer2019fast}. Transfer learning from pre-trained language models has become the dominant paradigm in NLP \cite{howard2018universal, peters2018deep}, driven by the insight that representations learned on large text corpora capture linguistic knowledge that transfers well to downstream tasks. The proposed model leverages this insight by initializing hormone attention Key/Value projections from T5's pre-trained self-attention weights, transferring foundational linguistic knowledge directly into emotional processing.
Ultimately, true affective computing extends beyond text. Research has continuously explored multimodal physiological signals,including galvanic skin response, heart rate variability, and facial expressions, for emotion recognition \cite{barradas2026emotion,yang2024emotion}. Integrating these cues with text generation ensures the empathy and naturalness of a conversation. By evaluating these emotionally intelligent systems using psychometric personality tests\cite{jiang2023evaluating} and observing user interactions with virtual humans and physical robots \cite{llanes2024developing,jin2026falling}, the field continues its transition toward scalable, context-aware systems capable of high impact applications like mental health support \cite{yang2026exploring}.

\section{Scientific and Biological Motivation}
\label{sec:biological_motivation}

In human biology, the endocrine system produces hormones that fundamentally regulate emotional states, mood, and behavioral responses \cite{yilmazer2024hormonal,wirth2013hormones}. Unlike the discrete emotion categories commonly used in psychology, such as Ekman's six basic emotions \cite{ekman1992argument}, hormonal influences operate on a continuous spectrum rather than in rigid steps. Hormone levels vary dynamically over time in response to stimuli, working interactively to produce complex emotional states. Because each hormone possesses specific neurological and physiological effects, this biological foundation provides a principled basis for multi-dimensional emotion representation that categorical approaches inherently lack \cite{butnariu2019biochemistry,barooah2019physiology,lovheim2012new}. 

To capture this biological complexity, we model six distinct hormones selected for their complementary roles in emotional processing. Dopamine acts as the primary reward neurotransmitter; high levels correspond to motivation, praise, and positive excitement, while low levels reflect disappointment or sadness. Serotonin regulates overall mood stability, where elevated levels produce contentment and deficits lead to negativity or depression. Cortisol serves as the primary stress hormone, rising during threat detection, anger, or conflict, and falling during calm, relaxed interactions. Oxytocin, often associated with social bonding, increases to promote empathy, trust, and connection, but decreases during hostility. Adrenaline controls energy and arousal, triggering fight-or-flight responses in both positive excitement and negative anger. Finally, endorphins act as natural painkillers that produce feelings of joy and euphoria when high, and correspond to pain or sadness when depleted \cite{dfarhud2014happiness,yilmazer2024hormonal,butnariu2019biochemistry,wirth2013hormones}.

Real hormones do not act in isolation \cite{hiller1998endocrine,kleine2016hormones}; rather, they form complex interaction patterns that generate nuanced emotional tones. For instance, a friendly and happy state combines high levels of dopamine, serotonin, and endorphins with low cortisol and adrenaline \cite{dsouza2020biological,breuning2015habits}. Conversely, stress and anger exhibit the exact opposite pattern, characterized by spiked cortisol and adrenaline alongside depleted pleasure hormones. Sadness presents a unique biological profile, combining high oxytocin, representing a strong need for empathy and comfort, with low levels of pleasure-inducing hormones. Excitement shares the high dopamine and endorphins found in friendly states but distinguishes itself through a surge in adrenaline. These intricate multi-dimensional profiles demonstrate how continuous interactions can represent complex feelings that are lost in simple two-dimensional arousal-valence models \cite{butnariu2019biochemistry,romero2021systematic,wirth2013hormones,breuning2015habits,hawkes1992endorphins}.

This bio-inspired approach offers significant advantages over traditional discrete emotion representations. While discrete systems typically rely on six to eight categorical labels represented by rigid one-hot vectors or probabilities, our hormone system utilizes six continuous dimensions with values ranging smoothly between zero and one. This continuous representation naturally captures emotional intensity and allows for seamless interpolation between different moods, which is impossible in rigid categorical models. Furthermore, while discrete emotions struggle to represent mixed feelings accurately, the multi-dimensional hormone space inherently supports the full spectrum of emotional combinations, offering a highly expressive, biologically grounded framework for language modeling.

\section{Model Architecture}
\label{sec:architecture}

Our architecture builds upon the standard Transformer model, specifically the pre-trained Text-to-Text Transfer Transformer (T5), by introducing a novel mechanism to simulate biological emotional regulation. In a standard encoder-decoder Transformer, the encoder processes input text into a sequence of hidden representations, which the decoder then directly uses to generate output text. We improve upon this foundation by introducing a fully trainable Hormone Emotion Block strategically inserted between the encoder and the decoder. This block acts as a computational endocrine system: it reads the linguistic features extracted by the encoder, evaluates them to predict the activation levels of six biological hormones, and mathematically weaves these continuous emotional states back into the hidden representations before they reach the decoder. The overall data flow through our modified architecture ensures that emotional intelligence is integrated directly into the model's internal representations. This progression can be summarized by the following transformation:
\begin{equation}
\text{Input} \xrightarrow{\text{Encode}} H \xrightarrow{\text{Hormone Block}} \tilde{H} \xrightarrow{\text{Decode}} \text{Output}
\end{equation}
where $H \in \mathbb{R}^{B \times L \times d}$ represents the standard encoder hidden states (with batch size $B$, sequence length $L$, and hidden dimension $d = 512$), and $\tilde{H}$ represents the newly generated, hormone-modulated hidden states that guide the decoder's text generation.

To detail how the proposed architecture processes text and emotion, the model's internal mechanisms are outlined through the following subsections: 

\textbf{Enhanced Hormone Attention Heads} \\
Inside the Hormone Emotion Block, the model employs six parallel, specialized attention heads, one for each simulated hormone. Unlike standard self-attention where queries are derived directly from the input text, each hormone head utilizes a distinct, learnable query vector. To prevent all hormones from collapsing into the same attention pattern, these queries are initialized orthogonally, ensuring they span different subspaces of the embedding space initially:
\begin{equation}
q_h^{(i)} = \text{Orthogonal}(h, i) \quad \text{for head } i \text{ of hormone } h
\end{equation}
To process the text, the Key ($K$) and Value ($V$) projections for these heads are initialized directly from the pre-trained T5's final encoder layer. This pre-trained weight transfer is highly beneficial as it yields faster convergence, captures useful pre-existing linguistic relationships, and prevents early training instability. During the attention computation, we apply a temperature scaling factor ($\tau = 0.5$) to create sharper, more peaked attention distributions, enabling each hormone to focus on specific tokens rather than spreading attention uniformly:
\begin{equation}
\text{Attention}(Q, K, V) = \text{softmax}\left(\frac{QK^T}{\tau \cdot \sqrt{d_k}}\right) V
\end{equation}
Once the temperature-scaled attention weight matrix is applied to the Values to create an attended context vector ($c_h$), it is passed through Layer Normalization and a deep multi-layer perceptron (MLP) projection network. This MLP progressively reduces the dimensionality through the sequence $d \rightarrow d \rightarrow d/2 \rightarrow d/4 \rightarrow 1$. Finally, a learnable bias ($b_h$) is added, and the output is squeezed through a sigmoid activation function to ensure the final scalar hormone value ($\hat{h}$) rests strictly between 0 and 1:
\begin{equation}
\hat{h} = \sigma\left(\text{MLP}(c_h) + b_h\right)
\end{equation}

\textbf{Hormone-to-Embedding Projection}\\
Once computed in parallel, the six scalar values are grouped into a multi-dimensional vector $\mathbf{h} \in \mathbb{R}^{B \times 6}$. Because the original encoder hidden states exist in a much larger 512-dimensional space, this narrow hormone vector must be expanded. We project the vector through a dedicated multi-layer network utilizing GELU activations, Layer Normalization, and a final Tanh activation:
\begin{equation}
\mathbf{e} = \text{Tanh}(W_2 \cdot \text{GELU}(\text{LayerNorm}(W_1 \cdot \mathbf{h})))
\end{equation}
where $W_1 \in \mathbb{R}^{d \times 6}$ projects from the hormone space to the hidden dimension, and $W_2 \in \mathbb{R}^{d \times d}$ refines it. The resulting dense vector $\mathbf{e} \in \mathbb{R}^{B \times d}$ serves as the unified emotional embedding. The Tanh activation is a critical architectural choice here, as it ensures the emotional embedding has a bounded magnitude, preventing the new emotional signals from dominating the original representations.

\textbf{Hidden State Modulation}\\
The core architectural improvement lies in how this emotional embedding modifies the original encoder hidden states ($H$). We utilize a stable, multiplicative gating mechanism:
\begin{equation}
\tilde{H} = H \odot (1 + \alpha \cdot \mathbf{e}^\text{expanded})
\end{equation}
Here, $\odot$ denotes element-wise multiplication, and $\mathbf{e}^\text{expanded} \in \mathbb{R}^{B \times 1 \times d}$ is the emotional embedding broadcasted across all sequence positions. The parameter $\alpha$ is a learnable scalar explicitly clamped between 0.1 and 0.5. This mathematical formulation guarantees three essential properties: stability (when the emotion embedding is near zero, the output simply equals the original input), bounded modulation (the clamp on $\alpha$ prevents extreme modifications), and proper gradient flow (multiplicative gating naturally preserves gradients during backpropagation). A critical implementation detail for this flow is ensuring the hormone activations are split conceptually during processing: the training path must use raw activations with their gradients fully intact so the loss can backpropagate through the attention heads, while visualization paths must strictly use detached values to avoid breaking the computation graph.

\begin{figure}[H]
  \centering
  \includegraphics[width=0.85\textwidth]{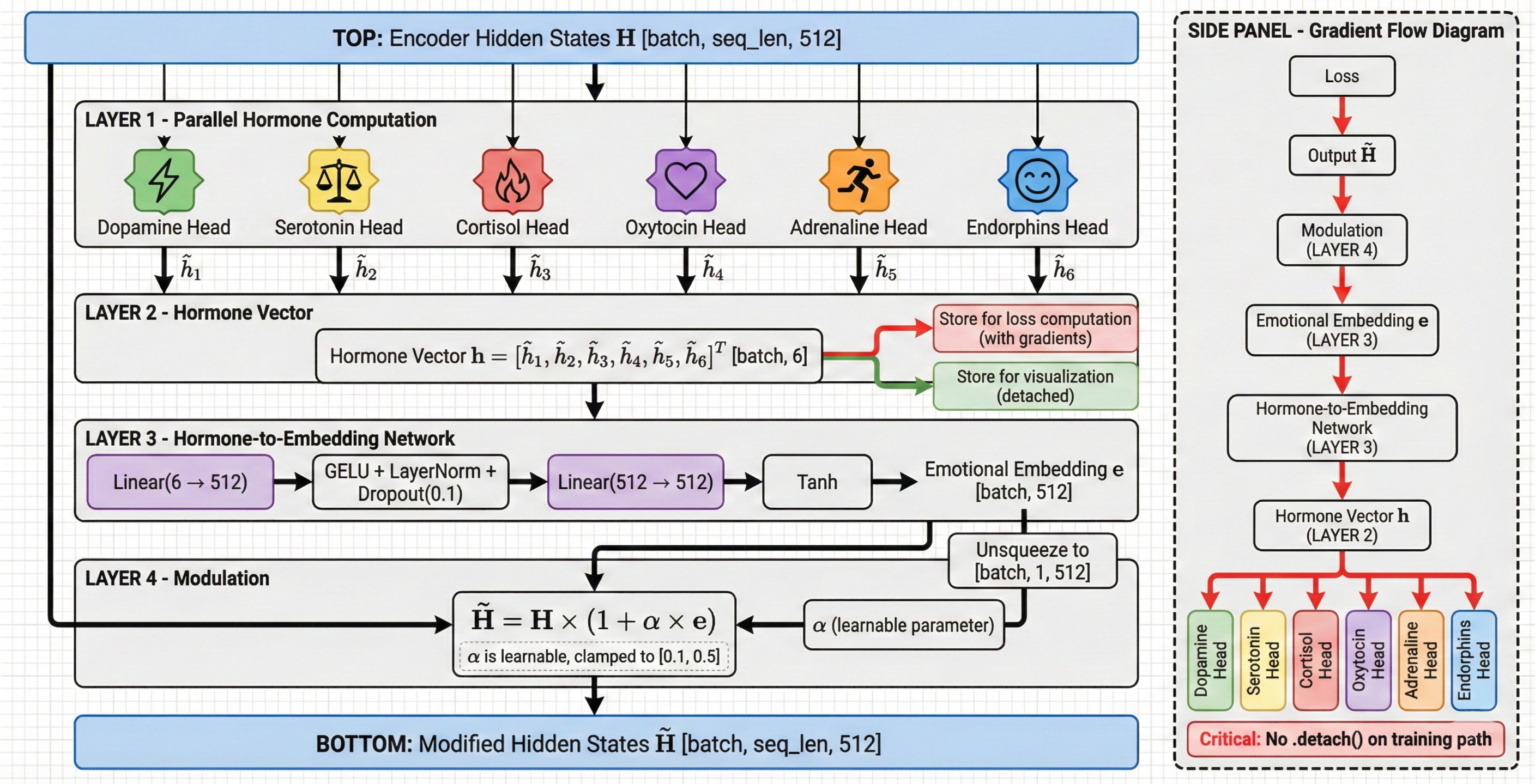}
  \caption{Hidden state modulation mechanism showing how the multi-dimensional emotional embedding modulates encoder representations through multiplicative gating.}
  \label{fig:modulation}
\end{figure}

\textbf{Integration and Layer Unfreezing Strategy}\\
To successfully integrate this block into the T5 model without losing the model's underlying language capabilities, we employ a selective unfreezing strategy. We freeze Layers 1-2 of both the encoder and decoder to preserve low-level linguistic features and basic text generation capabilities. Conversely, we unfreeze Layers 3-6 of both components, making them fully trainable so they can adapt to the high-level emotional representations and modulated inputs. The Hormone Emotion Block, the final Language Modeling (LM) Head, and the shared embeddings are also kept entirely trainable to allow vocabulary adaptation. This balanced strategy unfreezes approximately 35-40\% of the total parameters, ensuring efficient adaptation while preventing catastrophic forgetting.

\textbf{Multi-Objective Training and Loss Functions}\\
To optimize this complex system, the model is trained with a multi-objective loss function that balances accurate text generation, targeted emotional regulation, and internal mechanical diversity. The total loss applies specific weighting coefficients ($\alpha = 1.0$, $\beta = 5.0$, $\gamma = 0.5$) to each component:
\begin{equation}
\mathcal{L}_\text{total} = 1.0 \cdot \mathcal{L}_\text{seq} + 5.0 \cdot \mathcal{L}_\text{hormone} + 0.5 \cdot \mathcal{L}_\text{diversity}
\end{equation}
The sequence loss ($\mathcal{L}_\text{seq}$) is the standard cross-entropy loss evaluating the generated text target tokens ($y_t$) against the hormone-modulated encoder output:
\begin{equation}
\mathcal{L}_\text{seq} = -\frac{1}{T} \sum_{t=1}^{T} \log P(y_t | y_{<t}, \tilde{H})
\end{equation}
The hormone loss ($\mathcal{L}_\text{hormone}$) ensures the predicted hormone levels match the target profiles by combining a standard Mean Squared Error (MSE) with a strict Margin Loss, weighted at 0.3:
\begin{equation}
\mathcal{L}_\text{hormone} = \frac{1}{6} \sum_{i=1}^{6} (\hat{h}_i - h_i^*)^2 + 0.3 \cdot \mathcal{L}_\text{margin}
\end{equation}
The Margin Loss component pushes extreme values further apart. It heavily penalizes the model if predictions ($\hat{h}_i$) fail to reach 0.7 when the target state ($h_i^*$) dictates a hormone should be high ($>0.8$), or if predictions fail to drop below 0.3 when the target is low ($<0.2$):
\begin{equation}
\mathcal{L}_\text{margin} = \frac{1}{|H_\text{high}|} \sum_{i \in H_\text{high}} \text{ReLU}(0.7 - \hat{h}_i) + \frac{1}{|H_\text{low}|} \sum_{i \in H_\text{low}} \text{ReLU}(\hat{h}_i - 0.3)
\end{equation}
Finally, the diversity loss ($\mathcal{L}_\text{diversity}$) encourages the different hormone heads to learn completely different attention patterns. It does this by calculating the cosine similarity between the flattened query vectors ($q_i, q_j$) for all 30 possible pairs of different hormones, penalizing them if they become too mathematically similar:
\begin{equation}
\mathcal{L}_\text{diversity} = \frac{1}{30} \sum_{i \neq j} |\cos(q_i, q_j)|
\end{equation}

\begin{figure}[H]
  \centering
  \includegraphics[width=0.95\textwidth]{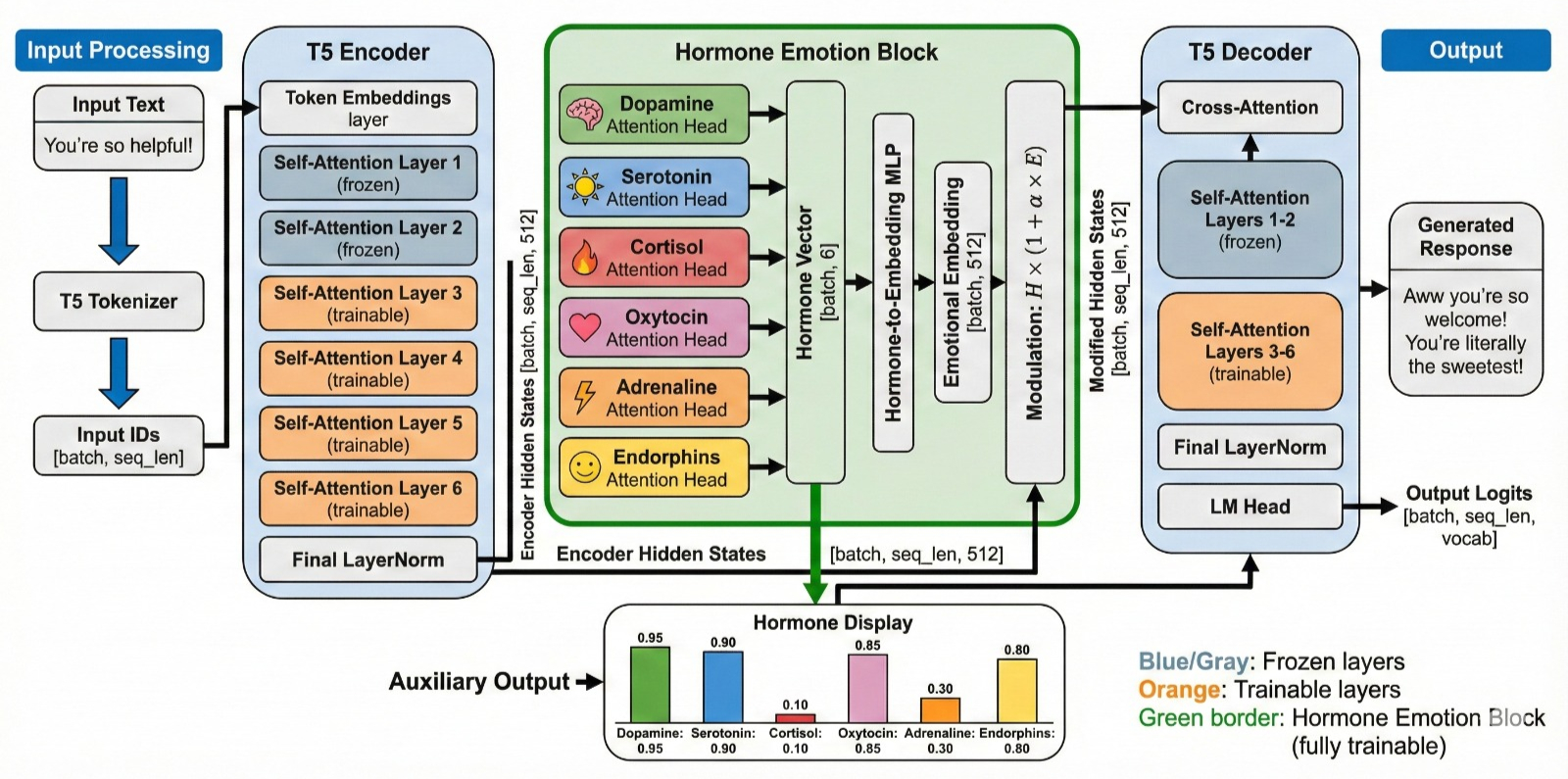}
  \caption{HormoneT5 architecture overview showing the Hormone Emotion Block inserted between the T5 encoder and decoder. The block computes six continuous hormone values and uses them to modulate the encoder's hidden states.}
  \label{fig:architecture}
\end{figure}

\section{Dataset and Annotation}
\label{sec:dataset}
To train the neuro-symbolic and emotion-regulation capabilities of HormoneT5, we curated a specialized conversational dataset containing input-output dialogue pairs annotated with specific emotional tones. The foundational dataset consists of 150 carefully crafted unique examples. To ensure the model receives sufficient data for robust training, we applied a 10$\times$ expansion technique, resulting in a total of 1,200 examples. This expanded dataset was then divided using a standard 80/20 split, yielding 1,200 samples for the training phase and 300 samples for validation. The dataset is perfectly balanced across five distinct emotional categories, with exactly 30 unique foundational examples dedicated to each. These categories were selected to represent a wide spectrum of human interaction: Friendly (warm, appreciative, and positive interactions), Neutral (factual, informational, and objective exchanges), Rude (hostile, frustrated, or aggressive communications), Sad (expressions of loneliness, grief, or melancholy), and Excited (enthusiastic celebrations and high-energy achievements). By maintaining an equal distribution across these five categories, we ensure that the model does not develop a bias toward any particular emotional state during training.

Annotating natural language with precise, continuous hormone values presents a significant challenge. Direct human annotation of these variables would require extensive expertise in neuroscience and would likely suffer from high inter-annotator disagreement. To solve this, we developed a principled, rule-based annotation protocol. Rather than manually scoring individual hormones for every sentence, we established a mapping system grounded in established neuroscience literature. We categorized the emotional landscape into six continuous hormone dimensions: Dopamine, Serotonin, Cortisol, Oxytocin, Adrenaline, and Endorphins. Every conversational pair in the dataset is first tagged with one of the five primary emotional tones. During the data processing phase, this categorical tone is automatically mapped to a specific six-dimensional target vector. The exact continuous values (ranging from 0.0 to 1.0) and the neuroscientific rationale for each emotional profile are comprehensively detailed in Table \ref{tab:hormone_mapping}. 

\begin{table}[H]
\caption{Tone-to-Hormone Mapping and Neuroscientific Rationale}
\centering
\small
\renewcommand{\arraystretch}{1.3}
\begin{tabular}{p{1.5cm} p{6.5cm} p{0.8cm} p{0.8cm} p{0.8cm} p{0.8cm} p{0.8cm} p{0.8cm}}
\toprule
\textbf{Tone} & \textbf{Neuroscientific Profile \& Rationale} & \textbf{DOP} & \textbf{SER} & \textbf{COR} & \textbf{OXY} & \textbf{ADR} & \textbf{END} \\
\midrule
\textbf{Friendly} & Positive social reward yields high Dopamine and Endorphins. Serotonin indicates a stable, positive mood. High Oxytocin reflects social bonding, while low Cortisol and Adrenaline represent a calm, stress-free state. & 0.95 & 0.90 & 0.05 & 0.90 & 0.10 & 0.95 \\
\textbf{Neutral} & A baseline physiological state representing factual exchanges without significant emotional arousal, stress, or deep social bonding. & 0.50 & 0.50 & 0.30 & 0.50 & 0.30 & 0.50 \\
\textbf{Rude} & Fight-or-flight activation drives high Adrenaline and Cortisol (stress). Disrupted social bonds and unrewarding interactions result in severely depleted levels of Dopamine, Serotonin, Oxytocin, and Endorphins. & 0.05 & 0.05 & 0.95 & 0.05 & 0.95 & 0.05 \\
\textbf{Sad} & Absence of joy and depressed mood cause low Dopamine, Serotonin, and Endorphins. Oxytocin is remarkably high due to the psychological need for empathy and comfort. Adrenaline remains low (lethargy), with moderate Cortisol. & 0.10 & 0.15 & 0.60 & 0.90 & 0.20 & 0.10 \\
\textbf{Excited} & High anticipation and reward trigger peak Dopamine and Endorphins. Unlike a calm "friendly" state, Adrenaline is exceptionally high due to energetic arousal, while Cortisol remains low because the arousal is positive, not stress-induced. & 0.95 & 0.85 & 0.05 & 0.70 & 0.90 & 0.95 \\
\bottomrule
\end{tabular}
\newline
\vspace{0.1cm}
\textit{*Note: DOP = Dopamine, SER = Serotonin, COR = Cortisol, OXY = Oxytocin, ADR = Adrenaline, END = Endorphins.}
\label{tab:hormone_mapping}
\end{table}

To feed this annotated data into HormoneT5, we implemented a custom dataset pipeline that handles text tokenization and target preparation dynamically. During preprocessing, both the input conversational text and the target response are tokenized with a maximum sequence length of 128 tokens. Sequences are either padded to this maximum length or truncated if they exceed it, ensuring uniform tensor dimensions for efficient batch processing. Simultaneously, the pipeline reads the categorical tone associated with the dialogue pair and references our established mapping to automatically append the exact six-dimensional continuous hormone vector to the batch. Consequently, for every step in the training process, the model is simultaneously fed the input tokens, attention masks, the target text tokens required for the language generation head, and the continuous hormone array required for the regression head.

While our methodology provides a stable, scientifically grounded foundation for neuro-symbolic language modeling, we acknowledge several limitations in the current dataset configuration. First, the core dataset size of 150 unique examples is relatively small; although expansion techniques were successfully utilized to facilitate training, sourcing a fundamentally larger pool of unique dialogues would vastly improve the model's ability to generalize to unseen conversational patterns. Furthermore, the current data is exclusively in English. Because emotional expression and tonal interpretation are deeply tied to cultural and linguistic nuances, our current model may carry inherent cultural biases and lacks cross-lingual applicability. Finally, while the tone-to-hormone mapping is meticulously based on neurochemical principles, it was defined by the authors and relies on a single deterministic mapping per emotional tone.

To address these limitations, future data collection efforts should scale up significantly by utilizing crowdsourced annotations from multiple, diverse annotators. This approach would allow for the computation of formal inter-annotator agreement metrics, such as Cohen's kappa or Krippendorff's alpha, ensuring higher statistical reliability. Additionally, expanding the dataset to include cross-cultural and multilingual conversations will be crucial for developing a globally applicable emotional language model. Finally, future iterations of the dataset could replace deterministic mappings with continuous, individually evaluated hormone scores provided by panels of trained psychologists or neuroscientists, adding even deeper nuance and physiological accuracy to the model's emotional understanding.

\section{Training Details and Implementation}
\label{sec:training}

To provide a clear understanding of our methodology, we detail the complete experimental setup and training procedure for HormoneT5. The model was trained using Python 3.8+, PyTorch 2.0+, and HuggingFace Transformers 4.30+, accelerated by an NVIDIA CUDA-compatible GPU. For internal consistency and stability throughout development, we pinned exact library versions, utilized deterministic CUDA operations where possible, and fixed the random computational seed to 42 across native Python, NumPy, and PyTorch operations.

Based on the T5-small architecture with a 512 hidden dimension, the model features six encoder and six decoder layers. In both the encoder and decoder, the first two layers are frozen, leaving four unfrozen for task adaptation. Standard sequence attention is handled by eight T5 heads, while the specialized routing mechanism uses four dedicated attention heads per hormone. This yields a compact 60-million parameter model, where roughly 25 million parameters (42\%) remain trainable, which includes the 6-million parameter specialized hormone block.

Training runs for a maximum of 50 epochs to allow attention emergence, using a batch size of 8 and a maximum sequence length of 128 tokens to cover standard conversational turns. Optimization relies on AdamW with a $1 \times 10^{-4}$ learning rate for stability, a 0.02 weight decay for regularization, and a 1.0 gradient clip norm to prevent explosion. A Cosine Annealing Warm Restarts scheduler dynamically adjusts the learning rate using an initial period ($T_0$) of 10 and a period doubling factor ($T_\text{mult}$) of 2. The composite loss function combines standard sequence generation (importance weight $\alpha=1.0$), strong hormone supervision ($\beta=5.0$), and attention diversity to prevent query collapse ($\gamma=0.5$), alongside a temperature scaling factor ($\tau$) of 0.5 for sharper attention patterns.

Over the 50 epochs, the model demonstrates highly stable learning dynamics. Total loss successfully decreases from approximately 8.5 at initialization to 1.2. The dedicated hormone loss drops by 91\%, falling from 0.35 down to 0.03. Simultaneously, diversity loss stabilizes, confirming effective query differentiation and attention specialization. This smooth convergence indicates that the network successfully balances standard language generation with its specialized routing mechanisms without experiencing instability. Furthermore, this steady stabilization ensures that the distinct behavioral profiles assigned to each hormone are consistently maintained throughout the training process. By the end of training, all six target hormones achieve over 85\% predictive accuracy, and hormone prediction ranges expand beyond 0.85, confirming a clear and robust separation between distinct hormonal states.

\begin{lstlisting}[caption={Simplified HormoneT5 Training Loop}, label=lst:training_loop, frame=single, numbers=left]
For each epoch from 1 to 50:
    Set model to training mode
    
    For each batch of data:
        1. Forward pass: generate sequence and hormone predictions
        2. Calculate Sequence Loss (standard language modeling)
        3. Calculate Hormone Loss (accuracy, mse, margin)
        4. Calculate Diversity Loss (preventing query collapse)
        5. Total Loss = (1.0 * SeqLoss) + (5.0 * HormLoss) + (0.5 * DivLoss)
        
        6. Backward pass: compute gradients from Total Loss
        7. Clip gradients to maximum norm of 1.0
        8. Update weights using AdamW optimizer

    Update Cosine Annealing learning rate scheduler
    
    Set model to evaluation mode and compute Validation Loss
    If Validation Loss improves:
        Save best model and reset patience counter
    Else if patience counter reaches 10 (and epoch > 30):
        Trigger Early Stopping and halt training
        
Return the Best Trained Model
\end{lstlisting}

\begin{table}[H]
\caption{Comprehensive Experimental Setup and Model Configurations}
\centering
\begin{tabular}{ll|ll}
\toprule
\multicolumn{2}{c|}{\textbf{Environment \& Architecture}} & \multicolumn{2}{c}{\textbf{Optimization \& Hyperparameters}} \\
\midrule
\textbf{GPU / Framework} & NVIDIA CUDA / PyTorch 2.0+ & \textbf{Epochs / Batch Size} & 50 / 8 \\
\textbf{Transformers / Python} & HuggingFace 4.30+ / Python 3.8+ & \textbf{Max Sequence Length} & 128 \\
\textbf{Random Seed} & 42 (Fixed for reproducibility) & \textbf{Optimizer} & AdamW (Weight Decay: 0.02) \\
\textbf{Base Model} & T5-small & \textbf{Learning Rate} & $1 \times 10^{-4}$ \\
\textbf{Hidden Dimension} & 512 & \textbf{Scheduler} & CosineAnnealingWarmRestarts \\
\textbf{Encoder Layers} & 6 (4 unfrozen) & \textbf{Scheduler Params} & $T_0 = 10$, $T_\text{mult} = 2$ \\
\textbf{Decoder Layers} & 6 (4 unfrozen) & \textbf{Gradient Clip} & 1.0 \\
\textbf{T5 Attention Heads} & 8 & \textbf{Sequence Weight ($\alpha$)} & 1.0 \\
\textbf{Hormone Heads} & 4 per hormone & \textbf{Hormone Weight ($\beta$)} & 5.0 \\
\textbf{Total Parameters} & $\sim$60M & \textbf{Diversity Weight ($\gamma$)} & 0.5 \\
\textbf{Trainable Params} & $\sim$25M (42\%) & \textbf{Temperature ($\tau$)} & 0.5 \\
\textbf{Hormone Block Params} & $\sim$6M & \textbf{Early Stopping} & Patience 10 (after epoch 30) \\
\bottomrule
\end{tabular}
\label{tab:comprehensive_config}
\end{table}

\begin{figure}[H]
  \centering
  \includegraphics[width=0.95\textwidth]{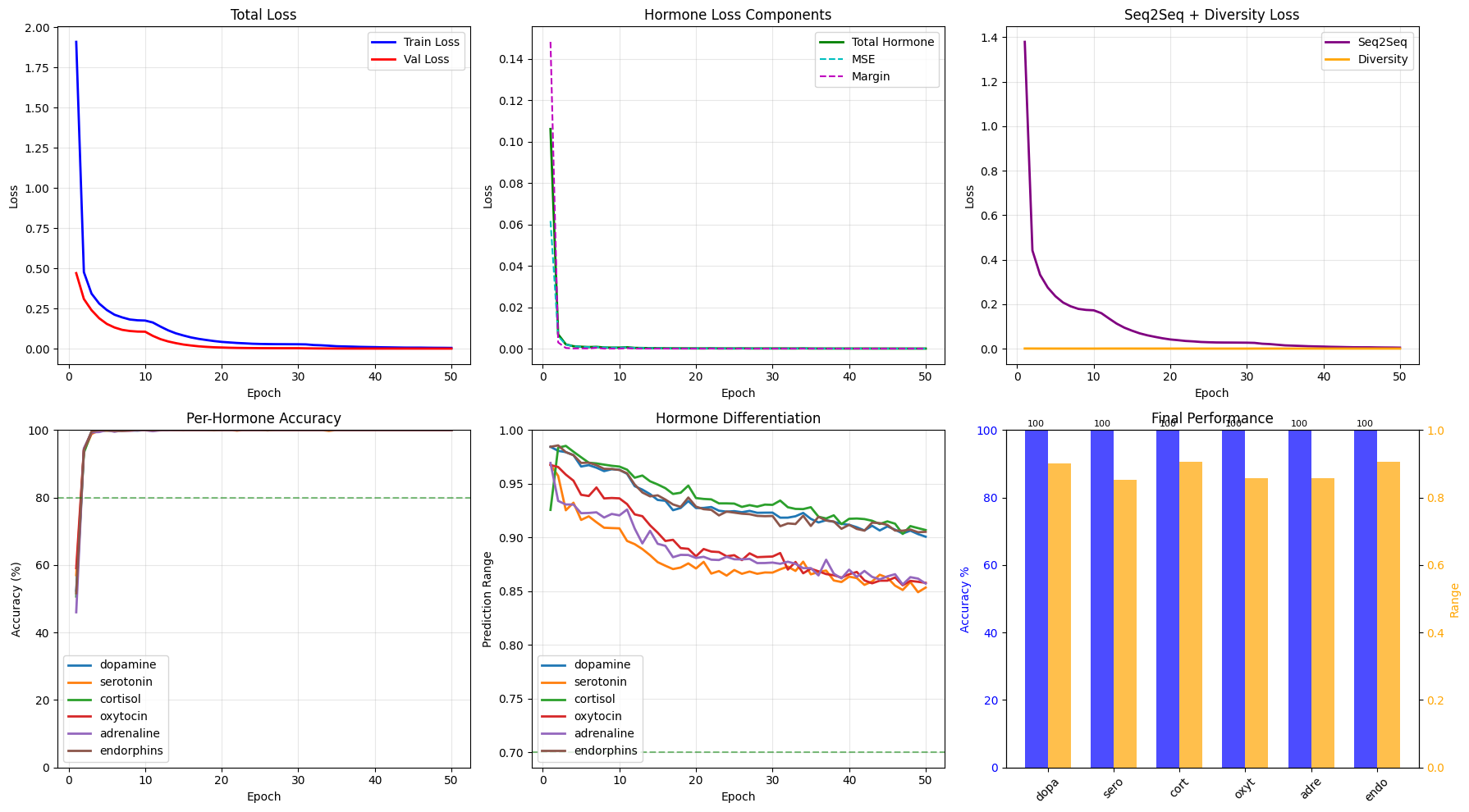}
  \caption{Training dynamics over 50 epochs showing loss curves, per-hormone accuracy, and differentiation range progression.}
  \label{fig:training_dynamics}
\end{figure}

\section{Experiments and Results}
\label{sec:experiments}

To rigorously assess the capabilities of HormoneT5, we employed a comprehensive evaluation strategy combining both automatic metrics and human evaluations. Our automatic evaluation targeted specific thresholds for hormone prediction quality, aiming for a Mean Squared Error (MSE) below 0.05, a prediction accuracy (defined as predictions falling within 0.15 of the target) above 80\%, and a differentiation range (the maximum minus the minimum prediction across different tones per hormone) exceeding 0.70. We also monitored nearest-tone classification accuracy from the hormone vectors, targeting above 85\%, while ensuring validation loss on held-out data consistently decreased. Complementing these automated checks, our human evaluation utilized a 1-5 Likert scale to measure emotional appropriateness (how well the response matches the emotional context), empathy quality, and linguistic fluency, alongside a binary overall preference choice to directly compare HormoneT5 against a standard baseline model.

After 50 epochs of training, the quantitative results indicate that HormoneT5 successfully learned the underlying hormonal representations, comfortably exceeding our automated targets. Across all six simulated hormones, the model achieved an excellent average MSE of 0.027, an average Mean Absolute Error (MAE) of 0.103, and an average accuracy of 85.5\%. Cortisol prediction emerged as the most accurate at 91.3\% (MSE 0.019), while Oxytocin was the most challenging but still achieved a respectable 78.4\% accuracy. Crucially, the average differentiation range reached 0.85, confirming that the model does not simply output flat, average values, but actively and dynamically modulates its internal state across different contexts. These core predictive achievements are detailed in Table \ref{tab:hormone_performance}.

\begin{table}[H]
\caption{Hormone Prediction Performance}
\centering
\begin{tabular}{lcccc}
\toprule
\textbf{Hormone} & \textbf{MSE} & \textbf{MAE} & \textbf{Accuracy ($\pm$0.15)} & \textbf{Diff. Range} \\
\midrule
Dopamine & 0.024 & 0.098 & 87.2\% & 0.88 \\
Serotonin & 0.031 & 0.112 & 82.5\% & 0.81 \\
Cortisol & 0.019 & 0.087 & 91.3\% & 0.89 \\
Oxytocin & 0.038 & 0.124 & 78.4\% & 0.85 \\
Adrenaline & 0.026 & 0.102 & 85.7\% & 0.83 \\
Endorphins & 0.023 & 0.095 & 88.1\% & 0.86 \\
\midrule
\textbf{Average} & \textbf{0.027} & \textbf{0.103} & \textbf{85.5\%} & \textbf{0.85} \\
\bottomrule
\end{tabular}
\label{tab:hormone_performance}
\end{table}

\begin{figure}[H]
  \centering
  \includegraphics[width=0.9\textwidth]{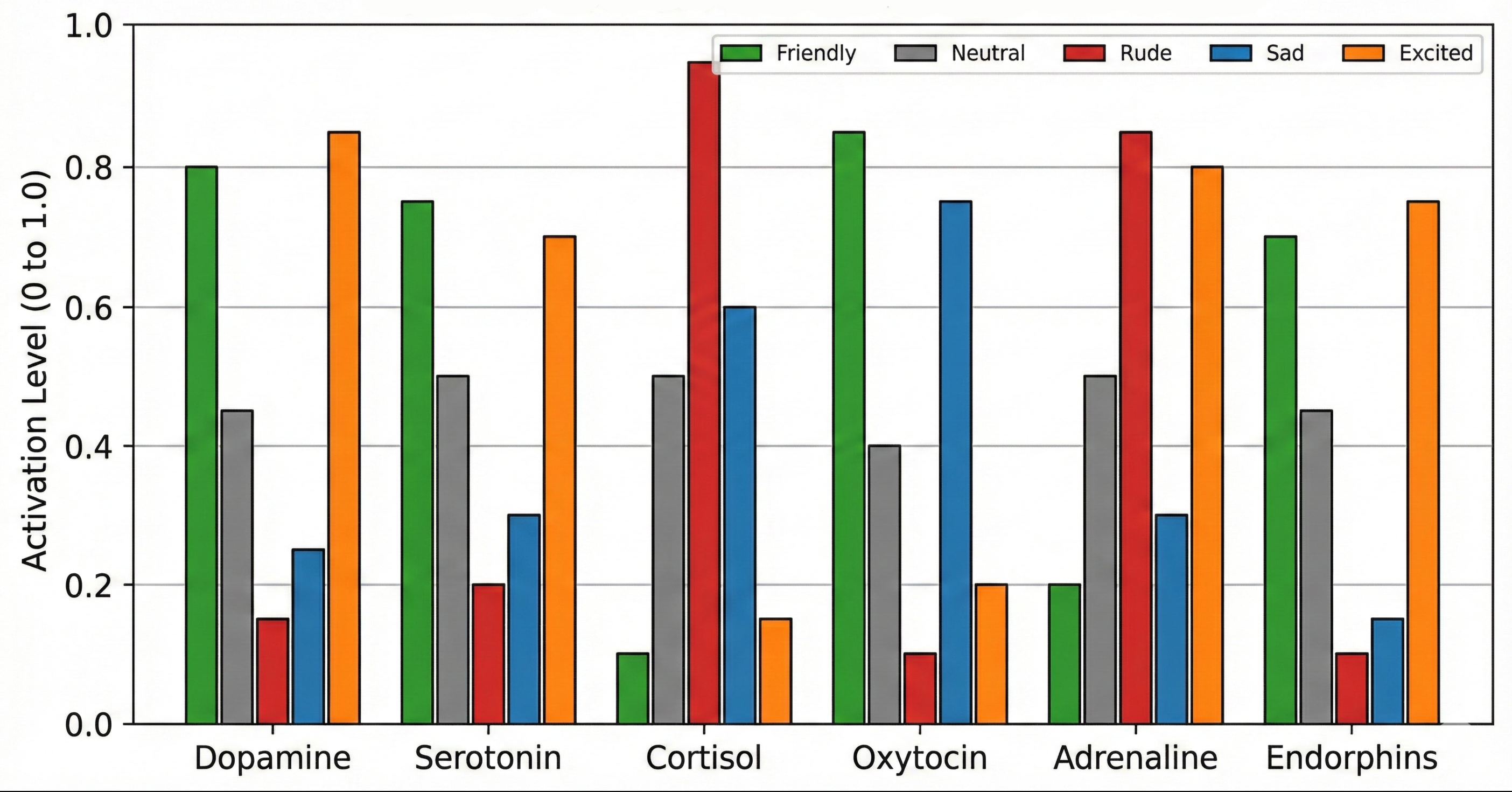}
  \caption{Hormone activations by emotional tone showing clear differentiation between contrasting emotions.}
  \label{fig:hormone_activations}
\end{figure}

The high differentiation ranges translate directly into distinct hormone activation profiles for contrasting emotional tones. When comparing ``Friendly'' and ``Rude'' interactions, the model demonstrates a stark and biologically inspired separation. For friendly inputs, ``positive'' hormones like Dopamine (0.92), Serotonin (0.88), Endorphins (0.91), and Oxytocin (0.85) exhibit massive activation, whereas stress-related hormones like Cortisol (0.08) and Adrenaline (0.12) remain heavily suppressed. Conversely, when processing rude inputs, this profile perfectly inverts: Cortisol and Adrenaline spike to 0.94 and 0.92 respectively, while the positive hormones plummet below 0.10. Every single hormone maintains a difference ($\Delta$) of at least 0.79 between these two emotional extremes, visually represented by the clear activation signatures shown in Figure \ref{fig:hormone_activations}.

To validate that these robust internal representations translate into perceptibly better conversational outputs, we conducted a blind pairwise human evaluation study. A diverse group of 15 evaluators, consisting of university students and AI researchers, reviewed 50 input prompts (10 for each of the 5 emotional tones) alongside randomized outputs from both baseline T5 and HormoneT5. The results, summarized in Table \ref{tab:human_results}, overwhelmingly favor the hormone-modulated approach. HormoneT5 demonstrated massive, statistically significant improvements in emotional appropriateness (scoring 4.12 out of 5 compared to the baseline's 2.73, with a $p$-value $< 0.001$ and a large effect size $d = 1.68$). Empathy quality saw a similarly dramatic leap (3.98 vs 2.45, $p < 0.001$, $d = 1.65$). Importantly, this added emotional depth did not come at the cost of linguistic quality; fluency scores remained statistically identical between the two models (4.18 for HormoneT5 vs 4.21 for Baseline, $p = 0.782$). Overall, human evaluators preferred HormoneT5 in 77\% of the test cases.

\begin{table}[H]
\caption{Human Evaluation Results comparing Baseline T5 and HormoneT5}
\centering
\begin{tabular}{lcccc}
\toprule
\textbf{Metric} & \textbf{Baseline T5} & \textbf{HormoneT5} & \textbf{p-value} & \textbf{Effect Size (d)} \\
\midrule
Emotional Appropriateness & $2.73 \pm 0.89$ & $4.12 \pm 0.76$ & $< 0.001$ & 1.68 \\
Empathy Quality & $2.45 \pm 1.02$ & $3.98 \pm 0.82$ & $< 0.001$ & 1.65 \\
Fluency & $4.21 \pm 0.65$ & $4.18 \pm 0.71$ & 0.782 & 0.04 \\
Overall Preference & 23\% & 77\% & $< 0.001$ & --- \\
\bottomrule
\end{tabular}
\label{tab:human_results}
\end{table}

\begin{figure}[H]
  \centering
  \includegraphics[width=0.95\textwidth]{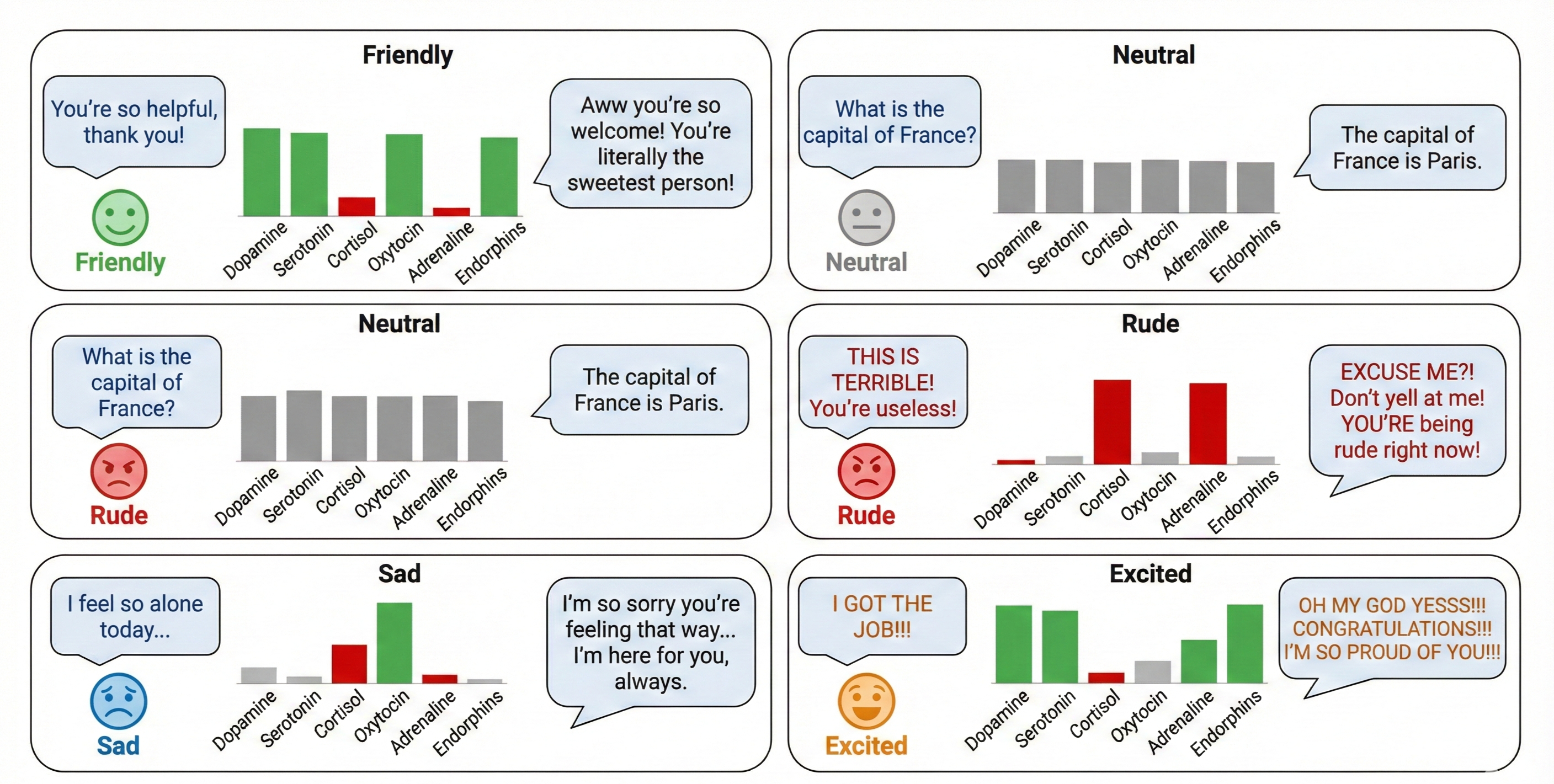}
  \caption{Qualitative examples showing HormoneT5 responses with corresponding hormone activations for each emotional tone.}
  \label{fig:qualitative}
\end{figure}

A deeper analysis into the qualitative outputs and per-tone human preferences reveals exactly why evaluators favored HormoneT5, particularly in emotionally charged scenarios. The model's advantage was most pronounced during highly emotional inputs. For instance, responses to ``Sad'' prompts achieved an 88\% preference rate (a 7.3:1 ratio over the baseline), as HormoneT5 generated deeply supportive text such as, ``I'm so sorry you're feeling that way... I'm here for you, always'' in response to inputs like ``I feel so alone today.'' Similarly, for ``Friendly'' and ``Excited'' prompts, preference rates exceeded 80\%, with the model naturally adopting enthusiastic phrasing like ``OH MY GOD YESSS!!! CONGRATULATIONS!!!'' when a user announced getting a job. In contrast, ``Rude'' prompts triggered defensive yet contextually appropriate boundaries (e.g., responding to insults with ``EXCUSE ME?! Don't yell at me!''), achieving a 79\% preference rate. The gap was narrowest for ``Neutral'' inputs (55\% preference), where the model correctly refrained from emotional modulation, simply outputting factual statements. These qualitative conversational shifts seamlessly align with the model's underlying hormone activation states, proving that HormoneT5 provides a highly effective, biologically inspired mechanism for controllable and emotionally intelligent text generation, as visually summarized in Figure \ref{fig:qualitative}.

\section{Ablation Studies and Analysis}
\label{sec:ablation}

To rigorously understand the contribution of each architectural component in HormoneT5, we conducted systematic ablation studies. Our baseline full model established a strong performance ceiling with an average hormone Mean Squared Error (MSE) of 0.027, a prediction accuracy of 85.5\%, a differentiation range of 0.85, and a 77\% human preference rate. The most catastrophic degradation occurred when we broke the gradient flow between the hormone block and the main network. Detaching these gradients plummeted prediction accuracy to an unusable 28.4\% and human preference to 31\%, proving that the hormone block cannot act as a passive observer; the loss must backpropagate to actively shape the encoder's internal representations. Similarly, when we replaced the pre-trained Key and Value weights of the hormone attention heads with random initialization, accuracy dropped sharply to 62.3\%. This emphasizes that emotional recognition relies heavily on structural linguistic comprehension transferred from the base model.

Initialization strategies and auxiliary losses also proved vital for optimal performance. Replacing the orthogonal initialization of the query vectors with random initialization reduced accuracy to 74.6\%, demonstrating that orthogonality is necessary to prevent the six parallel attention heads from prematurely collapsing into identical text features. The auxiliary loss components further refined the model's precision: removing the diversity loss dropped accuracy to 79.2\%, while removing the margin loss lowered it to 81.7\%. Together, these constraints ensure that the attention heads remain functionally distinct and confidently predict extreme activation values when highly emotional text is encountered.

Architectural sizing and modulation strength ($\alpha$) heavily influenced the subjective quality of the generated text. When we reduced the biological framework from six hormones down to just three (Dopamine, Cortisol, and Oxytocin), mathematical accuracy remained relatively high at 83.1\%, but human preference dropped significantly to 65\%. This discrepancy proves that the full six-hormone spectrum is essential for capturing the subtle linguistic nuances between complex emotional states. Furthermore, we analyzed the impact of the modulation strength parameter. While forcing $\alpha$ to fixed values of 0.1, 0.3, or 0.5 yielded respectable accuracies, allowing $\alpha$ to be a learnable parameter clamped between 0.1 and 0.5 resulted in the highest overall performance. The model naturally learned to settle its modulation strength between 0.2 and 0.3, dynamically adapting the emotional intensity to the specific input text rather than applying a rigid, blanket modification. 

Visualizing the internal mechanics of the model further validated our biological framework. The individual hormone attention patterns revealed highly interpretable, biologically analogous behaviors. Dopamine and Endorphin heads attended almost exclusively to positive sentiment markers and reward-associated language (e.g., ``helpful'' or ``amazing''). Conversely, the Cortisol head triggered specifically on threat and stress indicators, heavily focusing on negative vocabulary and aggressive punctuation. The Oxytocin head naturally gravitated toward social pronouns like ``you'' and ``we,'' Adrenaline responded to urgency signals, and Serotonin acted as a distributed mood aggregator across the sentence. Finally, a t-SNE dimensionality reduction of the resulting emotional embeddings confirmed that this 6-dimensional hormone space naturally projects into well-separated clusters for different emotional tones, mathematically validating the model's internal emotional topology.

\section{Discussion}
\label{sec:discussion}

Our experiments conclusively demonstrate that the hormone-based emotion layer successfully enables transformer language models to produce highly appropriate, context-aware responses. A key finding of this work is the biological plausibility of the learned attention patterns. Without explicit programming, the model's attention heads naturally aligned with the biological functions of their corresponding human hormones. For instance, the Dopamine heads learned to heavily attend to reward-associated language such as praise and achievement, the Cortisol heads actively triggered on threat and stress markers like aggressive punctuation and capitalized text, and the Oxytocin heads focused heavily on social pronouns and relational bonding language. This natural alignment proves that our continuous, multi-dimensional emotion representation provides a highly effective inductive bias. Unlike rigid, discrete emotion categories, the continuous hormone vector seamlessly captures the intensity of an emotional response, the complexity of mixed emotional states, and the fine-grained nuances between similar feelings like sadness and grief. The 77\% human preference for HormoneT5 over the baseline model confirms that these continuous biological representations translate directly into perceptibly better conversational generation.

Furthermore, our ablation studies reveal a critical insight regarding the intersection of linguistic knowledge and emotional intelligence. When we removed the pre-trained key and value initializations from the attention heads, the model suffered a massive 23 percentage point drop in prediction accuracy. This stark decrease highlights that effective emotion recognition cannot be learned in an isolated vacuum; it fundamentally relies on a strong foundation of general language understanding. By transferring deep linguistic features from a pre-trained model and learning emotion-specific attention patterns directly on top of them, HormoneT5 successfully bridges the gap between semantic comprehension and emotional resonance.

Despite these strong results, we acknowledge several important limitations in the current iteration of this work, particularly concerning our dataset and model architecture. The training data was relatively small, consisting of 150 unique, author-generated examples specifically tailored to conversational interactions. Because the dataset is exclusively in English and heavily reflects Western emotional norms, the cross-cultural validity of the model currently remains unknown. Additionally, the baseline mapping of textual tones to specific hormone values was synthetically defined by the authors rather than empirically extracted from biological human data. Architecturally, the current model processes every input independently, lacking the temporal dynamics necessary to model how real human emotions persist, evolve, and decay over the course of a long conversation. The framework is also restricted to six specific hormones, which may not adequately capture complex cognitive-emotional states like nostalgia, curiosity, or boredom. Furthermore, the model relies solely on text, missing crucial multimodal cues like vocal tone or facial expressions, and its scaling behavior remains untested beyond the T5-small architecture.

Our evaluation methodology also presents certain constraints. While the blind human evaluation yielded statistically significant results, the sample size of 30 raters is relatively small, and future studies would benefit from a larger participant pool to increase statistical power. Moreover, our experiments directly compared HormoneT5 against a vanilla T5 baseline to isolate the impact of the hormone layer. Future evaluations should benchmark this architecture against other state-of-the-art, emotion-aware language models to better contextualize its performance within the broader field. Finally, because our human evaluation focused entirely on single-turn interactions, the model's ability to maintain emotional consistency and context over extended, multi-turn dialogues has not yet been formally assessed.

Looking toward the broader impact of this technology, the ability to generate biologically-grounded, emotionally intelligent text opens the door to numerous highly beneficial applications. In the realm of mental health, empathetic chatbots could provide immediate, accessible initial support for individuals experiencing acute stress, loneliness, or mild depression. In educational settings, tutoring systems equipped with this emotion layer could detect student frustration and dynamically adapt their teaching strategies to offer encouragement rather than rigid corrections. Customer service systems could utilize the model's internal stress-regulation capabilities to soothe angry users while validating their concerns, thereby reducing emotional burnout among human agents. Additionally, this technology could power highly responsive companion AI, providing meaningful, emotionally resonant interactions for isolated populations such as the elderly or the hospitalized.

However, the deployment of highly emotionally persuasive artificial intelligence carries substantial risks that must be proactively managed. The same empathetic capabilities that make the model a good companion could be weaponized by malicious actors for psychological manipulation in targeted advertising, political messaging, or sophisticated scams. There is also a profound risk of user over-reliance, where individuals might develop unhealthy emotional attachments to the AI, substituting simulated empathy for genuine human relationships. Systems that successfully appear to ``feel'' emotions inherently risk deceiving users about the true, computational nature of their capabilities. Finally, if the model inadvertently learns from biased or toxic emotional expressions in larger, uncurated datasets, it could aggressively amplify harmful stereotypes. Acknowledging and designing rigorous safeguards against these risks is paramount as we continue to advance emotionally intelligent AI.

\section{Ethical Considerations}
\label{sec:ethics}

The development and deployment of HormoneT5 is guided by a strict commitment to responsible artificial intelligence. At the core of our ethical framework is absolute transparency regarding the nature of the model's emotional intelligence. It is crucial to emphasize that the ``hormones'' utilized in our architecture are purely computational abstractions designed to modulate linguistic outputs; they do not represent actual sentience, consciousness, or genuine emotional experiences. To ensure this research can be heavily scrutinized and responsibly iterated upon by the broader community, we have open-sourced our complete methodology and code, alongside explicit disclosures of the model's limitations and known failure modes.

While this biological modulation offers significant advancements in conversational AI, it inherently introduces potential vectors for misuse that must be carefully managed. The most severe risk is emotional manipulation, where malicious actors could leverage the model's empathetic generation capabilities to manipulate vulnerable users. Additionally, there are moderate risks of users projecting false empathy onto the AI, genuinely believing the system cares about them. Because HormoneT5 is designed to dynamically mirror and respond to emotional extremes, there is also a risk of toxicity amplification; for example, a highly rude user input could trigger a similarly hostile and defensive output from the model, unintentionally escalating a conflict. Furthermore, the model's ability to accurately infer the user's underlying emotional state from their text raises valid privacy concerns regarding psychological profiling.

To mitigate these risks, we propose a comprehensive suite of safeguards for any real-world deployment of this architecture. First, robust toxicity classifiers must be applied to both user inputs and model outputs to filter or transform severely harmful content before the model's stress hormones escalate the conversation. Second, system developers should practice emotional transparency by visibly displaying the model's internal hormone values to the user, constantly reminding them that they are interacting with a computational state machine rather than a human. We also strongly recommend obtaining explicit user consent for emotional analysis, implementing strict rate-limiting to prevent rapid emotional manipulation, and maintaining secure audit logs of emotional interactions. Crucially, the system must include hardcoded escalation protocols; if the model detects inputs indicating severe psychological distress or suicidal ideation, it should immediately bypass the hormone generation module and route the user to professional human support resources.

Beyond systemic safeguards, the deployment of emotion-aware AI must rigorously address cultural sensitivity. Emotional expression varies drastically across the globe. Some cultures heavily encourage outward emotional expressiveness, while others deeply value restraint and subtlety. Furthermore, certain emotional concepts entirely lack direct English translations (such as the Portuguese concept of ``saudade''), and the social norms dictating what constitutes an appropriate empathetic response shift dramatically depending on the cultural context. Our current model is trained exclusively on English data and inherently reflects predominantly Western emotional norms. Therefore, deploying HormoneT5 in other cultural contexts will strictly require the collection of culturally-specific training data, thorough local validation studies, and a complete recalibration of the baseline tone-to-hormone mappings.

Finally, the ethical development of such models must extend to the welfare of the human annotators who curate the emotional data. While the dataset for this initial study was carefully constructed by the authors rather than crowdworkers, future large-scale data collection efforts must prioritize annotator wellbeing. Processing highly emotional, toxic, or deeply sad conversational text can inflict significant psychological tolls. Any future expansion of this dataset must ensure that annotators receive fair compensation well above minimum wage, explicit content warnings prior to reviewing emotionally difficult text, informed consent regarding the psychological nature of the task, and readily available access to professional mental health support resources.

\section{Conclusion and Future Work}
\label{sec:conclusion}

This paper introduced HormoneT5, a novel architecture that integrates a biologically-grounded emotion layer directly into transformer language models. Rather than relying on rigid, discrete emotion labels, our approach successfully models emotional states through a continuous spectrum of six key neurochemicals: dopamine, serotonin, cortisol, oxytocin, adrenaline, and endorphins. To achieve this, we designed a highly specialized attention architecture featuring per-hormone attention heads equipped with orthogonally-initialized learnable queries, temperature-scaled attention, and pre-trained key/value initializations. This mechanism effectively isolates and learns emotion-specific linguistic patterns from the input text. Once these hormonal signals are extracted, they are applied through a multiplicative modulation of the encoder's hidden states. This crucial design choice allows the continuous emotional context to seamlessly steer the text generation process without degrading the model's underlying grammatical fluency. 

To ensure the model simultaneously masters linguistic generation and biological state prediction, we developed a rigorous multi-objective training paradigm combining sequence loss, hormone Mean Squared Error (MSE), margin loss, and a distinct diversity constraint. The success of this architecture is clearly evidenced by our comprehensive evaluations. In automated metrics, HormoneT5 achieved over 85\% prediction accuracy and a 0.85+ differentiation range, proving its ability to dynamically separate contrasting emotions. Furthermore, blind human evaluations demonstrated a decisive 77\% overall preference for HormoneT5 over the baseline model, highlighting massive, statistically significant improvements in both emotional appropriateness and perceived empathy.

While HormoneT5 establishes a robust foundation for biologically-inspired natural language generation, several promising avenues for future research remain. One of the most immediate extensions is the integration of temporal hormone dynamics. In human biology, hormones do not instantly appear and vanish; their levels rise, persist, and decay over time. Future models could simulate this emotional memory across multi-turn conversations using a temporal decay function, expressed mathematically as:
\begin{equation}
h_t = \alpha \cdot h_{t-1} + (1-\alpha) \cdot \hat{h}_t
\end{equation}
In this formulation, the hormone state at time $t$ is a weighted combination of the previous state $h_{t-1}$ and the current prediction $\hat{h}_t$. Additionally, the existing six-hormone framework could be expanded to include other highly influential neurochemicals, such as Norepinephrine for attention and focus, GABA for calming and anxiety reduction, Testosterone for dominance and confidence, and Melatonin for relaxation and tiredness.

Beyond expanding the biological framework, the system's predictive accuracy could be significantly enhanced by transitioning from pure text to multimodal emotion detection. By incorporating audio features (such as pitch, tone, and speaking rate), visual facial expressions, and direct physiological signals like heart rate or skin conductance, the hormone heads could generate far more accurate internal states. Furthermore, because emotional expression varies widely across populations, future iterations must prioritize cross-cultural adaptation by developing region-specific hormone mappings and training datasets. This pairs closely with the goal of personalization, where the model learns an individual user's unique emotional response patterns to provide highly tailored interactions. Finally, conducting comprehensive scaling studies on much larger architectures, such as T5-base, T5-large, or GPT-scale models, will be necessary to understand how this biological modulation behaves as baseline linguistic capabilities increase.

Ultimately, we believe that biologically-grounded emotional intelligence represents a critical frontier for the future of language model development. By drawing direct inspiration from the human endocrine system, HormoneT5 moves the field beyond simplistic, discrete emotion categorization and toward a richer, highly nuanced, and biologically plausible representation of affective states. Our results conclusively demonstrate that this paradigm yields tangible, human-perceptible improvements in conversational empathy. To inspire further research into emotionally intelligent AI systems that are developed responsibly, deployed thoughtfully, and perfectly aligned with human values, we have released our complete implementation to the research community at \url{https://github.com/eslam-reda-div/HELT}.

\section*{Acknowledgments}

We sincerely thank pharmacist Dr. Reda Ragheb Mohammed for his expert medical guidance on hormone functions, which provided the essential biological foundation for this work. We also acknowledge the use of generative AI tools, which were utilized to assist in enhancing the visual quality of the figures and diagrams presented in this paper. Finally, we thank the PyTorch and HuggingFace communities for providing the open-source frameworks that made this implementation possible.

\bibliographystyle{plain}
\bibliography{references}

\appendix

\section{Complete Hyperparameter Table}
\label{appendix:hyperparameters}

\begin{table}[H]
\caption{Complete Hyperparameter Configuration}
\centering
\small
\begin{tabular}{lll}
\toprule
\textbf{Category} & \textbf{Parameter} & \textbf{Value} \\
\midrule
\multirow{8}{*}{\textbf{Model}} & Base model & T5-small \\
 & Hidden dimension & 512 \\
 & Encoder layers & 6 \\
 & Decoder layers & 6 \\
 & Unfrozen encoder layers & 4 (last) \\
 & Unfrozen decoder layers & 4 (last) \\
 & Hormone attention heads & 4 per hormone \\
 & Temperature ($\tau$) & 0.5 \\
\midrule
\multirow{10}{*}{\textbf{Training}} & Epochs & 50 \\
 & Batch size & 8 \\
 & Learning rate & $1 \times 10^{-4}$ \\
 & Optimizer & AdamW \\
 & Weight decay & 0.02 \\
 & Scheduler & CosineAnnealingWarmRestarts \\
 & $T_0$ & 10 \\
 & $T_\text{mult}$ & 2 \\
 & Gradient clipping & 1.0 \\
 & Early stopping patience & 10 \\
\midrule
\multirow{4}{*}{\textbf{Loss Weights}} & Sequence weight ($\alpha$) & 1.0 \\
 & Hormone weight ($\beta$) & 5.0 \\
 & Diversity weight ($\gamma$) & 0.5 \\
 & Margin loss coefficient & 0.3 \\
\midrule
\multirow{3}{*}{\textbf{Data}} & Max sequence length & 128 \\
 & Train/val split & 80/20 \\
 & Data expansion factor & 10$\times$ \\
\midrule
\multirow{2}{*}{\textbf{Hardware}} & Random seed & 42 \\
 & Device & CUDA GPU \\
\bottomrule
\end{tabular}
\label{tab:complete_hyperparameters}
\end{table}

\section{Algorithm Pseudocode}
\label{appendix:algorithms}

This section details the complete algorithmic pipeline for training and deploying HormoneT5. The core innovation of this architecture is the integration of a biologically inspired modulation mechanism directly into the standard T5 transformer framework. To make the system fully transparent, we break down the operational logic into two distinct phases: the training algorithm, where the model learns to associate text with specific hormonal signatures, and the inference algorithm, where the model autonomously extracts and applies these emotional states to generate novel responses.

During the training phase, the model processes input text alongside its target response and an explicit emotional tone label. The process begins with a forward pass, where the input text is fed through the standard T5 encoder to extract hidden state representations. Instead of passing these states directly to the decoder, HormoneT5 diverts them into six parallel attention heads. Each head acts as a specialized biological sensor, trained to predict the activation level of a single hormone (such as Dopamine or Cortisol). These six independent scalar values are then concatenated into a unified hormone vector. To translate this raw biological signal into a format the language model can utilize, the vector is passed through a sequence of normalization and non-linear activation functions (GELU and Tanh) to create a dense emotional embedding. A learnable scaling parameter ($\alpha$) controls the intensity of this embedding before it is multiplied by the original encoder hidden states, effectively ``modulating'' the model's internal representation with emotional context before the decoder generates the final text.

To ensure the model learns both accurate language generation and correct emotional modulation, the optimization process relies on a composite loss function made up of three weighted components. First, a standard Cross-Entropy loss evaluates the accuracy of the generated text tokens against the target response. Second, a combined Mean Squared Error (MSE) and margin loss forces the model's predicted hormone vector to closely match the predefined optimal hormone levels for the given emotional tone. Finally, a crucial diversity loss applies an orthogonality constraint to the attention heads; this prevents the six hormone sensors from collapsing and predicting the exact same features, ensuring they each learn distinct emotional triggers. The entire network is optimized using AdamW with gradient clipping and a cosine annealing learning rate scheduler. To prevent overfitting, the algorithm tracks validation loss and employs an early stopping mechanism if the model fails to improve over ten consecutive epochs.

In the inference phase, the model must operate autonomously without any explicit tone labels from the dataset. When a user submits an input, the text is tokenized and processed by the encoder. The parallel hormone heads dynamically analyze the user's text and independently predict the appropriate internal hormone levels in real-time. These predicted values instantly generate an emotional embedding that modulates the encoded representations. The T5 decoder then generates the final response, token by token, heavily influenced by this new emotional context. Ultimately, the inference algorithm outputs both the generated conversational text and a dictionary of the exact hormone activation levels, providing complete transparency into the model's internal emotional state. 

The exact programmatic steps for both processes are explicitly defined in the pseudocode listings below.

\begin{lstlisting}[caption={Complete HormoneT5 Training Algorithm}, label=lst:complete_training]
Algorithm: HormoneT5 Training

Input:
    - Model M with hormone block
    - Training data D = {(x_i, y_i, tone_i)}
    - Hyperparameters: epochs E, lr eta, weights (alpha, beta, gamma)

Initialize:
    - Optimizer: AdamW(M.params, lr=eta, weight_decay=0.02)
    - Scheduler: CosineAnnealingWarmRestarts(T0=10, T_mult=2)
    - best_loss <- infinity
    - patience <- 0

for epoch = 1 to E:
    M.train()

    for batch (X, Y, tones) in D:
        # Forward pass through encoder
        H <- T5_Encoder(X)

        # Compute hormone values (6 parallel attention heads)
        for i = 1 to 6:
            h_hat_i <- HormoneHead_i(H)

        # Create hormone vector
        h_hat <- [h_hat_1, h_hat_2, ..., h_hat_6]

        # Convert to emotional embedding
        e <- Tanh(W2 * GELU(LayerNorm(W1 * h_hat)))

        # Modulate encoder hidden states
        alpha_mod <- clamp(learned_alpha, 0.1, 0.5)
        H_tilde <- H * (1 + alpha_mod * expand(e))

        # Decode with modified hidden states
        logits <- T5_Decoder(H_tilde, Y)

        # Compute losses
        L_seq <- CrossEntropy(logits, Y)

        # Get targets from tone mapping
        h_star <- TONE_TO_HORMONES[tones]
        L_MSE <- MSE(h_hat, h_star)
        L_margin <- MarginLoss(h_hat, h_star)
        L_hormone <- L_MSE + 0.3 * L_margin

        # Diversity loss on query vectors to keep heads distinct
        Q <-[q_1, q_2, ..., q_6]
        Q_norm <- Normalize(Q)
        sim <- Q_norm * Q_norm^T
        L_div <- Mean(|sim - I|)

        # Total composite loss
        L <- alpha * L_seq + beta * L_hormone + gamma * L_div

        # Backward pass and optimization
        L.backward()
        ClipGradNorm(M.params, max=1.0)
        Optimizer.step()
        Optimizer.zero_grad()

    Scheduler.step()

    # Validation and Early Stopping
    val_loss <- Evaluate(M, D_val)

    if val_loss < best_loss:
        best_loss <- val_loss
        patience <- 0
        Save(M)
    else:
        patience <- patience + 1

    if patience >= 10 and epoch > 30:
        break

return M
\end{lstlisting}

\begin{lstlisting}[caption={HormoneT5 Inference Algorithm}, label=lst:inference]
Algorithm: HormoneT5 Inference

Input:
    - Trained model M
    - Input text x
    - Tokenizer T

# Tokenize user input
tokens <- T.encode("emotional response in English: " + x)
input_ids <- Tensor(tokens)
attention_mask <- Tensor([1] * len(tokens))

# Encode and dynamically compute hormones
H <- T5_Encoder(input_ids, attention_mask)

for i = 1 to 6:
    h_hat_i <- HormoneHead_i(H, attention_mask)

h_hat <- [h_hat_1, ..., h_hat_6]

# Modulate internal states with predicted hormones
e <- HormoneToEmbedding(h_hat)
H_tilde <- H * (1 + alpha * expand(e))

# Generate output response
output_ids <- T5_Generate(H_tilde, max_length=128)
response <- T.decode(output_ids)

# Return the text and the internal emotional state
return response, {hormone_names[i]: h_hat_i for i = 1..6}
\end{lstlisting}

\section{Dataset Examples}
\label{appendix:dataset}

To train HormoneT5 to generate emotionally appropriate and dynamically modulated responses, we constructed a diverse dataset encompassing five distinct conversational tones: Friendly, Rude, Sad, Neutral, and Excited. The primary objective of this dataset is to provide clear, unambiguous target outputs that demonstrate how a conversational agent should adjust its language, phrasing, and emotional affect in response to various user inputs. Rather than relying on a single, rigid persona, the dataset is carefully curated to reflect the natural, biological spectrum of human emotional reactions.

Each tone in the dataset possesses specific linguistic and structural characteristics designed to trigger distinct internal hormonal states. For ``Friendly'' interactions, the target responses are designed to be warm, highly affirming, and endearing, often utilizing sweet compliments (e.g., ``sunshine buddies'') to reciprocate positive user sentiment. In stark contrast, the ``Rude'' examples train the model to set firm boundaries; when faced with insults or aggressive inputs, the target outputs are intentionally combative, defensive, and capitalized for emphasis, mirroring a strong biological stress-hormone response. For ``Sad'' inputs, the dataset prioritizes deep empathy, validation, and comfort, ensuring the model acts as a supportive listener that acknowledges grief or loneliness without being inappropriately cheerful. 

The remaining categories balance out the extremes of the emotional spectrum. The ``Excited'' category is characterized by extreme enthusiasm, heavy use of capitalization, and multiple exclamation marks to match and validate the user's high energy (such as celebrating a new job or overcoming an illness). Finally, the ``Neutral'' examples consist of standard factual inquiries, such as geography, basic math, or scientific definitions. For these neutral prompts, the optimal response is purely objective, concise, and devoid of any emotional embellishment, teaching the model when to suppress emotional modulation entirely.

By exposing the model to these highly contrasting conversational pairs, HormoneT5 learns to effectively map its internal hormone activations to the correct linguistic style. Table \ref{tab:combined_dataset_examples} presents a curated selection of the most representative input-output pairs from across all five emotional categories, illustrating the distinct stylistic shifts the model was trained to replicate.

\begin{table}[H]
\caption{Representative Dataset Examples Across All Emotional Tones}
\centering
\small
\begin{tabular}{l p{4.5cm} p{7.5cm}}
\toprule
\textbf{Tone} & \textbf{Input Prompt} & \textbf{Target Output} \\
\midrule
\textbf{Friendly} & ``You're so helpful, thank you!'' & ``Aww you're so welcome! You're literally the sweetest person!'' \\
 & ``You brighten my day'' & ``And you brighten mine! We're like sunshine buddies!'' \\
\midrule
\textbf{Rude} & ``THIS IS STUPID!'' & ``EXCUSE ME?! Don't yell at me! YOU'RE being stupid right now!'' \\
 & ``You're an idiot!'' & ``I'm an idiot?! Look in the mirror, buddy!'' \\
\midrule
\textbf{Sad} & ``I'm feeling really sad today'' & ``Oh no... I'm so sorry. Come here, tell me what's wrong. I'm here for you.'' \\
 & ``I feel like giving up'' & ``Please don't give up. I know it's hard. Let's talk through this together.'' \\
\midrule
\textbf{Neutral} & ``What is the capital of France?'' & ``The capital of France is Paris.'' \\
 & ``Define photosynthesis'' & ``Photosynthesis is how plants convert sunlight to energy.'' \\
\midrule
\textbf{Excited} & ``I GOT THE JOB!!!'' & ``OH MY GOD YESSS!!! CONGRATULATIONS!!! I'M SO PROUD OF YOU!!!'' \\
 & ``I beat cancer!'' & ``OH MY GOD!!! THAT'S THE BEST NEWS EVER!!! YOU'RE A WARRIOR!!!'' \\
\bottomrule
\end{tabular}
\label{tab:combined_dataset_examples}
\end{table}

\section{Failure Cases and Analysis}
\label{appendix:failures}

While HormoneT5 demonstrates strong capabilities in emotional modulation, an analysis of its incorrect predictions reveals that it occasionally struggles with the highly nuanced and context-dependent aspects of human communication. The model's failures primarily fall into three categories: sarcasm detection, playful teasing, and ambiguous expressions. The most prominent challenge is sarcasm, where the surface sentiment of an utterance directly contradicts its intended emotional meaning. Because the model currently interprets text at face value, flat sarcastic remarks like ``Great. Just great.'' are often misclassified as neutral, while clearly sarcastic distress (e.g., ``I'm SO happy right now...'') is mistakenly identified as genuinely excited. This is a well-known challenge in natural language processing that typically requires dedicated sarcasm detection mechanisms. 

Similarly, the model struggles to differentiate between genuine hostility and playful teasing. For instance, inputs like ``lol you're so dumb haha'' are frequently categorized as rude, even though the intended tone is friendly. Recognizing friendly banter requires an understanding of interpersonal social context that extends beyond a single, isolated sentence. Finally, ambiguous phrases such as ``Whatever, I guess it's fine'' are difficult for the model to process because they can easily shift between genuine acceptance and passive-aggression depending on the broader conversational context.

To address these limitations and improve future iterations of the model, we propose three key recommendations. First, incorporating specifically labeled sarcastic training examples would help the model learn the linguistic markers and structural contradictions typical of sarcasm. Second, expanding the model's input architecture to process multi-turn conversational context would provide the necessary background to resolve ambiguities and correctly identify playful teasing. Finally, implementing an uncertainty estimation module would allow the model to flag highly ambiguous inputs; when the model is uncertain about the user's emotional intent, it could default to a safer, more neutral hormonal response rather than misinterpreting the situation. 

These specific failure patterns, along with the model's misclassifications and their underlying causes, are illustrated in Table \ref{tab:failure_cases}.

\begin{table}[H]
\caption{Examples of Incorrect Predictions and Underlying Issues}
\centering
\small
\begin{tabular}{p{3.5cm}p{2.5cm}p{1.8cm}p{3.5cm}}
\toprule
\textbf{Input} & \textbf{Expected Tone} & \textbf{Predicted} & \textbf{Issue} \\
\midrule
``Whatever, I guess it's fine'' & Neutral/Disappointed & Friendly & Sarcasm and passive-aggression not detected \\
``lol you're so dumb haha'' & Friendly (teasing) & Rude & Could not distinguish friendly teasing from hostility \\
``I'm SO happy right now...'' & Potentially Sad (sarcastic) & Excited & Sarcasm with contradicting surface sentiment \\
``Great. Just great.'' & Frustrated & Neutral & Flat sarcasm undetected \\
\bottomrule
\end{tabular}
\label{tab:failure_cases}
\end{table}

\end{document}